\documentclass{article}
\usepackage[utf8]{inputenc}
\usepackage[numbers]{natbib}
\usepackage{graphicx}
\usepackage{xcolor}
\usepackage{caption}
\usepackage{subcaption}
\usepackage{url}

\title{\textbf{Sensors, Safety Models and A System-Level Approach to Safe and Scalable Automated Vehicles}}
\author{Jack Weast, Intel}
\date{\today}

\begin{document}

\maketitle

\section*{Abstract}
When considering the accuracy of sensors in an automated vehicle (AV), it is not sufficient to evaluate the performance of any given sensor in isolation. Rather, the performance of any individual sensor must be considered in the context of the overall system design. Techniques like redundancy and different sensing modalities can reduce the chances of a sensing failure. Additionally, the use of safety models is essential to understanding whether any particular sensing failure is relevant. Only when the entire system design is taken into account can one properly understand the meaning of safety-relevant sensing failures in an AV. 

In this paper, we will consider what should actually constitute a sensing failure, how safety models play an important role in mitigating potential failures, how a system-level approach to safety will deliver a safe and scalable AV, and what an acceptable sensing failure rate should be considering the full picture of an AV's architecture.

\section*{What is a sensing failure in an AV?}
Numerous research projects attempt to prove the accuracy (\cite{2010_geiger}, \cite{mu2019mnist}, \cite{lin2014microsoft}) of object detection and classification algorithms. These projects often feature leader boards where researchers compare the accuracy of a novel new algorithm for a given data set. Popular belief is that a single instance of a sensing algorithm (and associated sensing modality) is only suitable to use in the context of a safety-critical application like an AV when that algorithm is proven to have near zero failures on a given data set.

\begin{figure}[h!]
\centering
\includegraphics[scale=0.2]{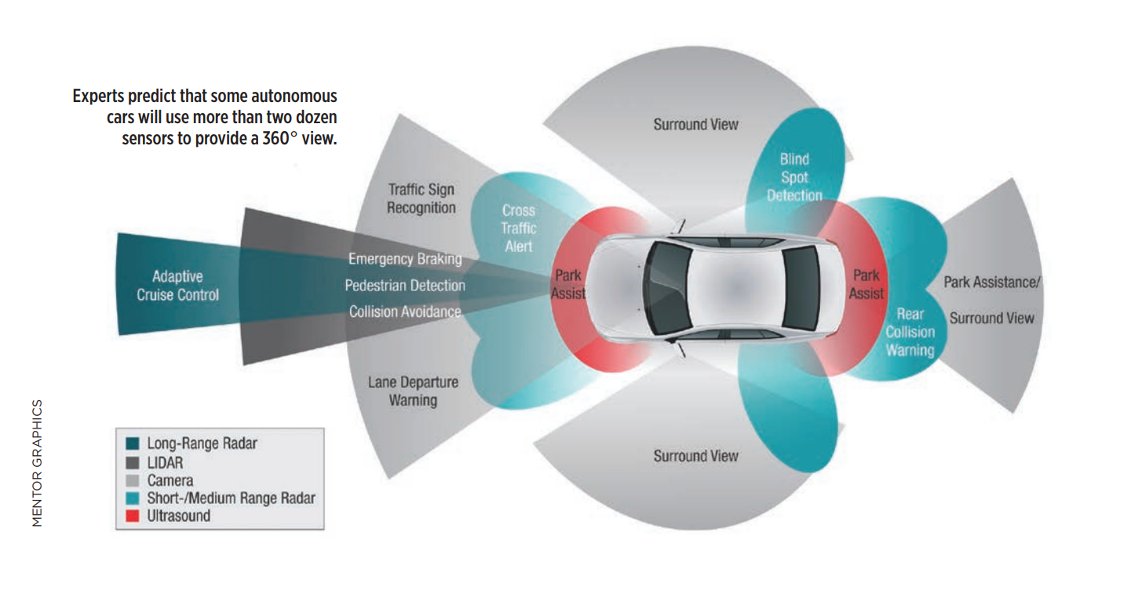}
\caption{Example of a typical Sensing System of an AV \cite{sae_sensors}}
\label{fig:av_example}
\end{figure}

Yet these efforts have very little correlation to an AV in the real world.  No AV operates with a single sensor or sensing algorithm in isolation; AVs rely on multiple sensors of many different types.  As shown in Figure \ref{fig:av_example}, cameras, radars, lidars and other sensing modalities provide overlapping 360-degree coverage along with a high-definition map to contribute to an accurate creation of a world model so that when the AV makes decisions, it is doing so with an accurate representation of the real world.

The traditional approach to creating a world model is to combine camera, radar and lidar into a single sensing channel. This approach works well so long as all sensors continue to operate properly. While there may be redundancy in terms of overlapping fields of view between sensing types, there is no independence between the sensing subsystems. This presents a problem if any one of the sensing types did not work properly. What if the front facing camera has mud on the lens? What if the radar is confused by large metal plates on the road?  Is a single sensing channel strong enough to still interpret the environment when there is missing information coming from faulty/unavailable sensors?  Because the sensing system was designed to rely on all three sensors to create an accurate world model, likely not.  This is why reliance on a single sensing channel to accurately perceive the world is insufficient and unsafe. 

\begin{figure}[h!]
\centering
    \begin{subfigure}[b]{0.7\textwidth}
        \centering
        \includegraphics[width=\textwidth]{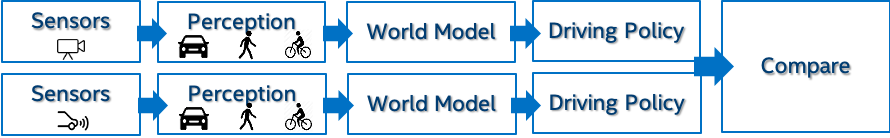}
        \caption{Redundant and Independent Sensing system}
    \end{subfigure}
    \hfill
    \begin{subfigure}[b]{0.7\textwidth}
        \centering
        \includegraphics[width=\textwidth]{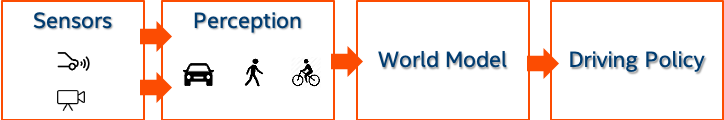}
        \caption{Redundant Sensor system}
    \end{subfigure}
\caption{Redundant and Independent vs. Redundant Sensor Systems}
\label{fig:redundant_independent_sys}
\end{figure}

Instead, what if an AV could operate safely with cameras alone? What if another AV could operate safely with just radar and lidar? What would happen then if those two independent sensing subsystems were combined into a single AV? This approach provides both redundancy and independence of sensing, so that if any one channel is temporarily unavailable, the other channel can continue to safely operate the AV. The result is a more robust system and, therefore, a safer AV (see Figure \ref{fig:redundant_independent_sys}).

With two independent sensing subsystems, both would have to fail at exactly the same time in order for there to be a sensing failure that may (or may not, as we will see later) lead to a safety incident. Let us say that the probability of failure for either sensing channel is p. As they are independent, the probability of them both failing at the same time is p2. Consider that the human level probability of a crash is approximately 2x10-5. While not likely, for purposes of consideration assume a worst-case scenario that a sensing failure of any kind would cause a crash. If each of the redundant and independent sensing subsystems were only as good as a human, the overall probability that the two systems will fail together at exactly the same time and lead to a crash is $4.10^{-10}$, which is 50,000 times better than a human driver.

But what of our original question?  What constitutes a sensing failure in an AV? Sensing failures traditionally fall into two categories:

\begin{itemize}
    \item False Positive – A sensor believes an object is present when it is not
    \item False Negative – A sensor believes there is no object present when in fact there is
\end{itemize}

False positives may prompt the AV to brake or swerve unnecessarily, thereby producing an uncomfortable ride. But so long as other drivers are maintaining safe distances, doing so will not cause a crash.

False negatives then are the primary concern for safety. But are all false negatives created equal? If a sensor fails to see an object that is a mile away, is that a failure that should count against the sensing subsystem? If a sensor accurately identifies an object in 59 out of 60 frames, should that be considered a failure?

We assert that sensing failures only matter if they could lead to a safety incident (e.g. a crash).  Meaning that if the sensing failure has no impact on safety of the driving task, then it should not be considered a failure. But how to define “safety” in this context? To answer this question, we must turn our attention to the role of safety models in an AV.

\section*{What does safety mean for an AV?}
Human drivers understand that the act of driving is inherently risky. The safest human driver would be the person that never drives. So too for an AV – the safest AV in the world would be the one that never leaves the garage. 

The human transportation network is designed such that informed risks must be taken in order to get where we need to go. Necessary maneuvers to create space when cutting into traffic are often provably unsafe, yet humans do this all the time and mostly without crashing. Human drivers often follow too close and cannot guarantee they would avoid crashing into the car they are following if it were to suddenly brake at maximum theoretical braking force.

The challenge then is how to define safety in such a way so the AV drives more safely than humans while still taking a certain amount of risk in order to navigate in the real world.  With such a definition, we can then consider what impact a sensing failure may have on the actual safety of the AV as it operates in the real world.

Traditional industry approaches to defining safety rely on statistical arguments around things like number of miles driven. Claims of having driven a million miles, or even 10 million miles, are insufficient measures of safety, primarily because little is known about the time, location or properties of the miles driven. What's more, gathering sufficient statistical evidence to make an actual claim regarding safety would require 10's of billions of miles driven \cite{kalra2016driving}, which is infeasible. As a result, statistical arguments cannot be used to understand or argue the future safety of an AV.

Other arguments to define safety rely on a perfect application of the traffic rules. It sounds simple enough; if an AV drives exactly according to the traffic rules, it should be safe. But consider the simple example of a metered intersection where the AV has a green light, meaning it has the right of way. Yet, there is a vehicle crossing the intersection as a result of that driver running the red light. According to the rules of the road, the AV has the right of way and so has every right to proceed and crash into the other vehicle. And the fault for the accident would lie with the other driver for running the red light. But clearly this was an avoidable accident and, in this case, strictly following the rules of the road does not result in an AV that drives safely.

\begin{figure}[h!]
\centering
\includegraphics[scale=0.15]{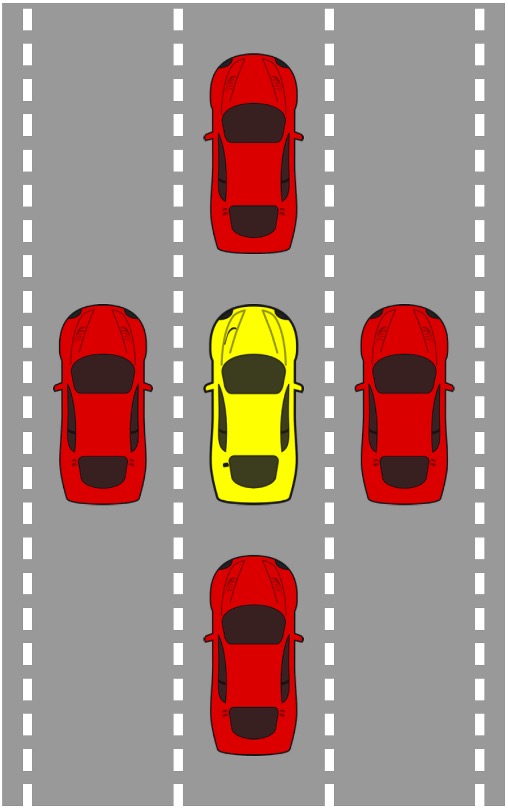}
\caption{Nothing can be done to guarantee safety}
\label{fig:ego_surrounded}
\end{figure}

Also common are ethical approaches to defining safety. It's popular among ethicists \cite{luetge2017german} to say that the AV's goal is to avoid collisions at all costs. Consider then a three-lane highway during rush hour and the AV is in the middle lane surrounded by other vehicles, as illustrated in Figure \ref{fig:ego_surrounded}. Is there anything the AV can do to avoid a crash if one of the adjacent drivers is intent on crashing into the AV? In order to fulfill this overriding ethical principle, the AV would never be able to drive in the middle lane on a highway, let alone any scenario where an adjacent car could turn into the AV if the AV has no other escape route. Such an AV may fulfill the ethical principle, but would no longer be a useful vehicle that anyone would want to ride in.

How then to define what it means for an AV to drive safely so everyone can understand whether a sensing failure would lead to a situation that would lead to a safety incident?

Consider this question in the context of human drivers. The best human drivers can drive their entire lives without ever having caused an accident. Why? Was it just luck? Or is there something that AV's can learn from the best human drivers in order to drive safely?

The human transportation system is based on rules. The first kind, \textit{Explicit} rules, are well understood by both humans and machines alike. Speed limits, stop signs, double yellow lines, etc., all represent explicit traffic rules that must be followed. But there is another set of rules.  These are not obvious and are culturally relative. These \textit{Implicit} rules cover things like keep a safe distance or drive cautiously if pedestrians could be occluded. But what constitutes a “safe distance”?  How would an AV know if it's driving “cautiously” or not? For that matter, how would a human?

\textit{Implicit} rules are in fact generalized behavioral characteristics that are often cultural. What it means drive cautiously in China may mean something completely different in the United States. A safe distance may mean one thing in Germany but something entirely different in Italy. Yet, it is these Implicit traffic rules that best represent to humans what it means to “drive safely.” The challenge is to take these implicit, culturally sensitive rules and formalize them in such a way that an AV, driven by a computer, can then “drive safely”.

That is exactly what is captured in the Responsibility-Sensitive Safety (RSS) model \cite{shalev2017formal}  – an open and transparent, technology neutral, formal model for the implicit rules of driving. The RSS model can be used to define what it means to drive safely, balancing safety and usefulness in a way that is culturally sensitive to the norms of safe driving wherever the AV is operating.

The Responsibility-Sensitive Safety Model
RSS is comprised of 5 basic rules:
\begin{enumerate}
    \item Do not hit the car in front
    \item Do not cut in recklessly
    \item Right of way is given, not taken
    \item Be cautious in areas with limited visibility
    \item If the vehicle can avoid a crash without causing another one, it must do so
\end{enumerate}

RSS captures these rules in a set of mathematical formulas and logic that is transparent and mathematically verifiable. The resulting “safety model” enables the AV to drive carefully enough so that it will not be the cause of an accident, and cautiously enough so that the AV can compensate for the reasonable and foreseeable mistakes of others. RSS can be utilized within an AV to provide a check on the probabilistic nature of the AV's decision-making artificial intelligence (AI) algorithms. It can also be used externally to assess the performance of AVs in the real world to ensure they are driving safely according to the model.

There are three elements to the RSS safety model. The first is a formalization of what it means to drive safely; in practice this is understood as a safe longitudinal and lateral distance around the vehicle. Just like humans subconsciously maintain a safety envelope around the car when driving, so does the AV using RSS. When that safety envelope is compromised, we are in a \textit{dangerous situation}, which is the second element. The moment preceding a \textit{dangerous situation} is called the \textit{danger threshold}, and it is at this point that the AV must perform an action to restore the safe distances that have now been compromised. This action, the third element of the model, called a \textit{proper response}, defines how the AV should respond to the dangerous situation that was caused by the actions of other drivers.

\section*{RSS in Practice}

\begin{figure}[h!]
\centering
\includegraphics[scale=0.7]{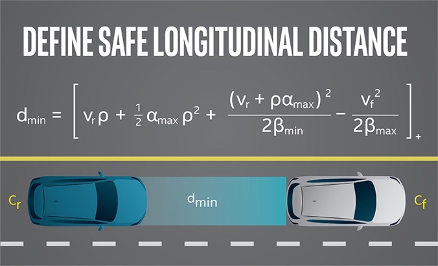}
\caption{RSS Rule \#1}
\label{fig:rss_rule1}
\end{figure}

The first RSS rule states that the AV should not hit the car in front and uses a formula to determine how much distance is needed between the AV and the car it is following so that the AV can stop if the car in front were to brake suddenly. With Newtonian physics as a guide, we define a minimum distance formula based on the current velocity of the rear vehicle $v_r$, the response time of the AV $\rho$, the possible maximum acceleration by the AV over the response time $\alpha_{max}$, the velocity of the front vehicle $v_f$, and the assumed reasonable and foreseeable maximum deceleration of the front vehicle $\beta_{max}$, as shown in Figure \ref{fig:rss_rule1}.

With the safe longitudinal distance defined, we can identify what constitutes a \textit{dangerous situation} in a car following scenario, which would happen if the front vehicle were to brake, triggering the \textit{danger threshold} at which the AV would perform the \textit{proper response} and restore the minimum safe distance.

\begin{figure}[h!]
\centering
\includegraphics[scale=0.7]{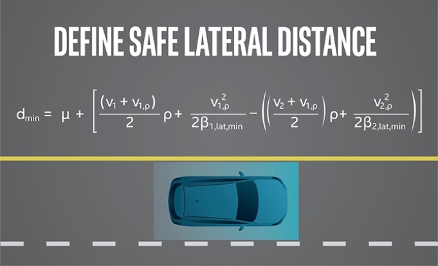}
\caption{RSS Rule \#2}
\label{fig:rss_rule2}
\end{figure}

Using the same physics-based approach, we can also define a minimum safe distance for lateral scenarios which is the second rule of the RSS safety model: Do not cut in recklessly, illustrated in Figure \ref{fig:rss_rule2}.

For these two rules, an important concept exists that traces back to the fact that driving safely is inherently cultural and involves a balance of risk and usefulness. This is embodied in parameters that represent a reasonable and foreseeable assumption about the worst-case behavior of other drivers. Human drivers make assumptions about other vehicles all the time. 

We assume when following another vehicle that it will not brake with the force of a race car, but rather at a reasonable braking force like a human. We also assume when driving next to another vehicle in the same direction of travel that the car next to us isn't going to suddenly turn into our lane at an aggressively high velocity.

While it is true that the car we are following may decelerate, and the car next to us may inadvertently turn into our lane, these are normal, reasonable and foreseeable circumstances that if the AV performs the proper response, the AV can safely avoid a collision.

But if the AV was required to guarantee that it would never crash into the car it is following – in any situation, anytime, anywhere – then the AV would have no choice but to assume a global maximum for the theoretical worst-case braking capability of any vehicle on the road anywhere in the world. In practice this would mean the car with the most significant braking capability (such as a race car) would set the following distance for the AV all of the time as that is the only way to ensure there would never be a crash from behind. Such a vehicle would maintain extremely long following distances. Although certainly safe, it would not be very useful, would worsen traffic flow, and would likely annoy other drivers on the road.

This is why an AV must be able to make reasonable and foreseeable assumptions about the behaviors of other road users. These assumptions, embodied in parameters in the RSS formulas, are what sets the culturally sensitive balance between safety and usefulness. RSS parameters can by adjusted and tuned to conform with societal expectations of where the AV is being deployed.

\begin{figure}[h!]
\centering
\includegraphics[scale=0.7]{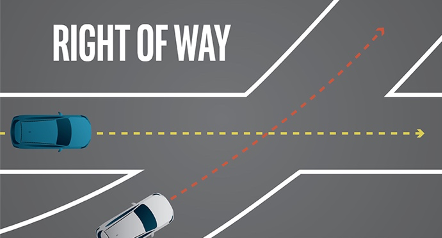}
\caption{RSS Rule \#3: Right of Way is Given, not Taken}
\label{fig:rss_rule3}
\end{figure}

The third rule of RSS states simply that Right of Way is given, not taken. There are many traffic scenarios where just because an explicit signal says a particular vehicle has right of way does not mean it should take it. In Figure \ref{fig:rss_rule3}, the blue car has the right of way. But if it's clear that the silver car is not going to stop (even when the explicit rules require it to do so), the safe move is for the blue car to yield, as right of way is given and not taken, just as a human would do when driving safely.

\begin{figure}[h!]
\centering
\includegraphics[scale=0.7]{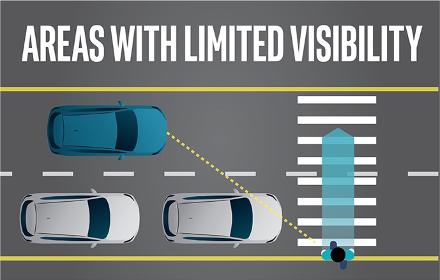}
\caption{RSS Rule \#4: Be Cautious in Areas of Occlusion}
\label{fig:rss_rule4}
\end{figure}

The fourth rule of RSS (Figure \ref{fig:rss_rule4}) states that the AV should be cautious in areas with limited visibility. There are many scenarios where pedestrians may be occluded behind other vehicles or perhaps a car driving around a bend is occluded by a large building. In both scenarios the AV needs to be able to make reasonable and foreseeable assumptions about the worst-case behavior of other agents. For a pedestrian that may be behind a moving truck, what is their maximum possible speed? For a vehicle coming around the bend, how fast should the AV expect they could be driving? Making the correct assumptions is what will allow the AV to drive safely based on local norms and behaviors.

\begin{figure}[h!]
\centering
\includegraphics[scale=0.7]{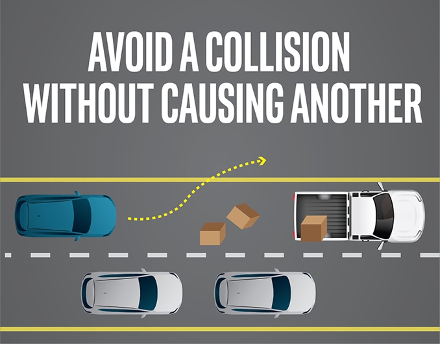}
\caption{RSS Rule \#5: Avoiding a safety incident by performing another safe maneuver}
\label{fig:rss_rule5}
\end{figure}

Finally, the fifth rule of RSS (Figure \ref{fig:rss_rule5}) states that if the AV can avoid a crash without causing another one, then it must do so. This is just common sense and something that all humans would do as well when driving safely. As illustrated in Figure 8, if it were possible to safely change lanes to evade some boxes that were falling out of the back of a truck without causing a new crash, then the AV should do so. This rule also helps an AV drive safely in the case where other road users do not behave within the reasonable and foreseeable assumptions (e.g. a car that suddenly brakes faster than expected). In this case the AV would be expected to perform an evasive maneuver in order to restore a safe state without causing another accident. 

In summary, RSS provides a clear, easy to understand model for industry, the public and governments to define what it means for an AV to drive safely.

\section*{Putting it All Together}
Since RSS is a physics-based model, it requires an accurate understanding of the position and velocity of other road users. When no other vehicle failures are happening, and all other road users are accurately perceived and operating within the bounds of their reasonable and foreseeable worst-case behavior defined by RSS, then there should be no accidents. Based on these assumptions, this means only sensing failures could possibly lead to a safety incident.

This brings us back to the original question: What exactly does it mean to have a sensing failure? Perhaps a better question is, what are the relevant sensing failures when it comes to the safety of an AV?

With RSS as a guide, an AV's ability to drive safely can now be understood by considering the impact of a sensing failure as it relates to the AV's ability to comply with and perform the rules of the safety model, and therefore, drive safely.

For example, if a sensing failure resulted in an AV not performing a proper response to a dangerous situation, then that could lead to a safety incident. By contrast, if a sensing system failed to recognize an object 50 meters off the side of the road, no different action would have been required according to the safety model and so this sensing mistake does not impact the AV's ability to comply with it, meaning it would not have affected the AV's ability to drive safely. The result is a significant reduction in the situations where a sensing failure could lead to a safety incident.

RSS can further mitigate the effects of statistically known sensing failures while also improving comfort. It does this by adding distance to the AV's calculations just in case the sensors are slightly off.  Just because the AV can safely drive within a few centimeters of a parked car doesn't mean that it should, else it be ‘too close for comfort' from a human perspective.

RSS can also help mitigate against transient sensing failures. For example, if the classification of an object changes throughout its tracking life, the parameters specified in the model can provide guidance on the reasonable and foreseeable worst case, so that no matter what the object is and how it was classified by the system for a short period of time, the AV can safely navigate around it.

In short, sensing failures that do not affect the AV's ability to comply with the safety model should not be deemed sensing failures.

\section*{What Sensing Failure Rate is Acceptable?}
In the technology industry, the Mean Time Between Failures (MTBF) is a measure of how reliable the system is as it provides the mean failure rate over time.

Having a self-driving system that the public can trust requires a significant increase with respect to driving safety compared to human drivers. As noted previously, the probability of a human driver causing an injury accident in one hour of driving is $p = 2.10^{-5}$, resulting in a MTBF ($1/p$) of about 50,000 hours of driving. The AV should not be only as good as a human, it should be better than a human driver. If an AV has a MTBF of $10^6$ hours, which is 20 times better than a human driver, but is part of a fleet of 1 million AVs, this still equates to one accident every hour, which might be unacceptable to society.

So, the challenge then is to identify the acceptable MTBF for an AV sensing system.  A MTBF of $10^7$ hours would mean that a sensing failure that could lead to the AV’s inability to comply with the safety model should not happen more than once in 1 million hours of driving. In order to statistically prove such a claim, an AV would need to drive 10 million hours without a safety incident, which if an AV drove 30 miles per hour, is 30 Billion miles of driving, clearly infeasible with even the largest fleet of AVs.

An alternative approach to achieving this same level of MTBF would be to leverage the system-level redundancy and independence of the sensing system described earlier.  This makes it possible to achieve a MTBF goal of $10^7$ hours by reducing the MTBF goal of each sensing system to $10^4$ hours.

In this way, the MTBF of a camera-only subsystem that would lead to an inability to comply with the RSS model is $10^4$ hours and the MTBF of a radar and lidar sensor subsystem that would lead to an inability to comply with the RSS model is also $10^4$ hours.  As noted before, the probability of two independent sensing systems failing at exactly the same time is the product of their independent failure rates, thus in this case the MTBF would be $10^8$, delivering an AV that is 10,000 times safer than a human driver, as well as exceeding the MTBF goal of $10^7$.

Achieving an independent sensing subsystem that achieves a $10^4$ MTBF is still not an easy task, as that is equivalent to driving 2 hours a day for 10 years without a sensing failure that would lead to a safety incident. However, with sufficiently mature sensing algorithms, proving such a failure rate is in fact achievable; with a fleet of 100 vehicles operating concurrently, that proof can be achieved in a few months, a significant reduction in the validation burden.

On the other hand, an AV with a single, even redundant but not independent, sensing system would have to achieve a MTBF of $10^8$ on a single channel, which would be equivalent to driving 2 hours a day for 10,000 years without ever making a software update, as doing so would require restarting the validation effort. Even with the largest AV fleet imaginable operating in parallel, it would be inconceivable that anyone would be able to complete such an activity and be able to make such a claim. 

\section*{Conclusion}
When considering sensing failures in AV's, it is not sufficient to consider the performance of a single sensor in isolation. One must consider system level redundancies and independence of the sensing systems. Further, not all sensing failures are meaningful, only those that would lead to an inability to comply with a verifiable safety model. The use of RSS provides important clarity on what sensing failures are safety relevant so that one may better understand how to understand and assess sensing failures, while importantly providing a definition of what it means to drive safely so that industry the public and government can have an open and honest conversation about the safety of AV's.

\bibliographystyle{unsrtnat}
\bibliography{references}

\section*{About the Author}
Jack Weast is a Senior Principal Engineer at Intel and Vice President for Automated Vehicle Standards at Mobileye. In this role, Jack leads a global team working on AV safety technology and the related standards that will be needed to understand what it means for an AV to drive safely. 

\end{document}